\documentclass[runningheads]{llncs}
\usepackage{graphicx}
\usepackage{color}
\usepackage{xcolor}
\usepackage{multirow}
\usepackage{algorithm}
\usepackage{algorithmic}
\usepackage{makecell}
\usepackage{stfloats}
\usepackage{amsmath,amssymb,amsfonts}
\usepackage{pifont}
\usepackage{arydshln}
\usepackage{xcolor}
\usepackage{marvosym}
\usepackage{bm}
\definecolor{mygland}{RGB}{80,255,200}
\usepackage{hyperref}

\begin{document}
\title{Morphology-inspired Unsupervised Gland Segmentation via Selective Semantic Grouping}
\titlerunning{Morphology-inspired UGS via Selective Semantic Grouping}
%
\author{Qixiang Zhang\inst{1} \and
Yi Li\inst{1} \and Cheng Xue\inst{2} \and
Xiaomeng Li\inst{1}\textsuperscript{(\Letter)}}

\authorrunning{Qixiang Zhang et al.}

\authorrunning{Q. Zhang et al.}

\institute{Department of Electronic and Computer Engineering, The Hong Kong University
of Science and Technology, Hong Kong, China\\ \email{eexmli@ust.hk}
 \and School of Computer Science and Engineering, Southeast University, Nanjing, China
}

%
\maketitle              
\begin{abstract}
Designing deep learning algorithms for gland segmentation is crucial for automatic cancer diagnosis and prognosis. However, the expensive annotation cost hinders the development and application of this technology. In this paper, we make a first attempt to explore a deep learning method for unsupervised gland segmentation, where \emph{no manual annotations are required}. Existing unsupervised semantic segmentation methods encounter a huge challenge on gland images. \textbf{They either over-segment a gland into many fractions or under-segment the gland regions by confusing many of them with the background.} To overcome this challenge, our key insight is to introduce an empirical cue about gland morphology as extra knowledge to guide the segmentation process. To this end, we propose a novel Morphology-inspired method via Selective Semantic Grouping. We first leverage the empirical cue to selectively mine out proposals for gland sub-regions with variant appearances. Then, a Morphology-aware Semantic Grouping module is employed to summarize the overall information about glands by explicitly grouping the semantics of their sub-region proposals. In this way, the final segmentation network could learn comprehensive knowledge about glands and produce well-delineated and complete predictions. We conduct experiments on the GlaS dataset and the CRAG dataset. Our method exceeds the second-best counterpart by over \textbf{10.56\%} at mIOU.

\end{abstract}
\section{Introduction}
Accurate gland segmentation from whole slide images (WSIs) plays a crucial role in the diagnosis and prognosis of cancer, as the morphological features of glands can provide valuable information regarding tumor aggressiveness~\cite{colorectal}. 
With the emergence of deep learning (DL), there has been a growing interest in developing DL-based methods for semantic-level~\cite{MSFCN,concl,ASM} and instance-level~\cite{glas,DCAN,FRN,MIDL,GlandInstanceSeg,TANET} gland segmentation. However, such methods typically rely on large-scale annotated image datasets, which usually require significant effort and expertise from pathologists and can be prohibitively expensive~\cite{histopathologySurvey}.

\begin{figure}[t]
\centering
\includegraphics[width=\textwidth]{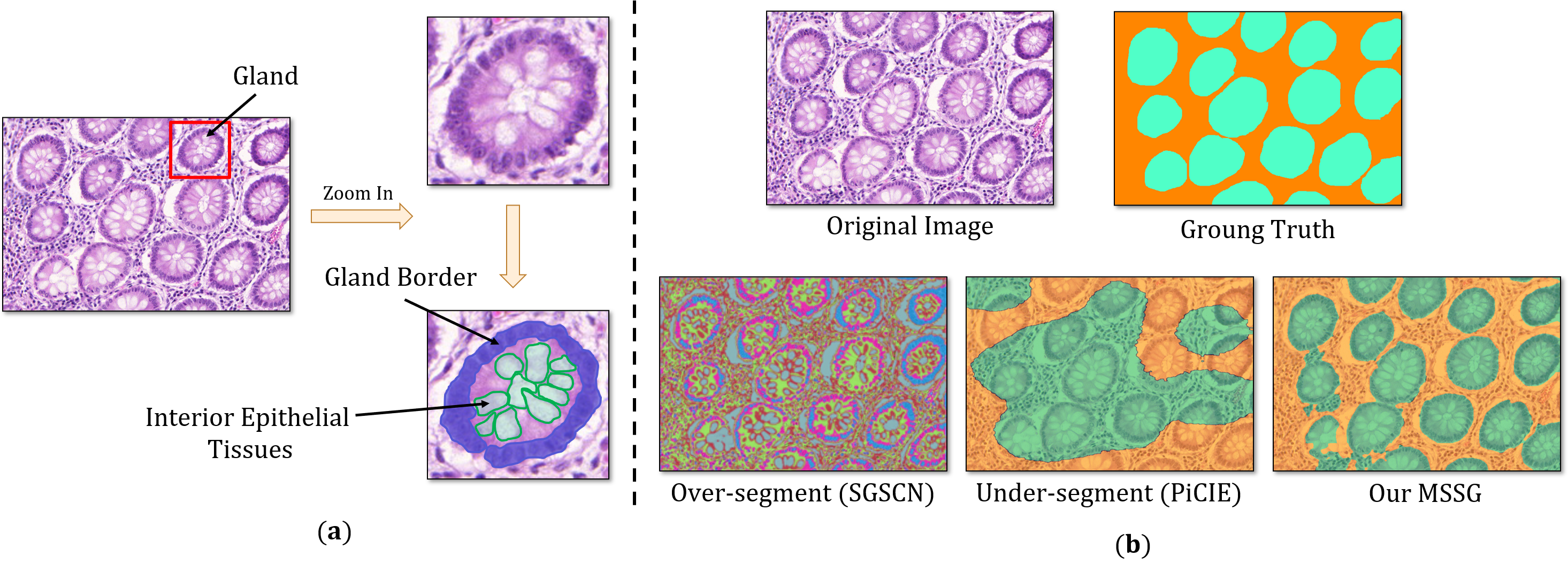}
\caption{(a): Example of a gland and its gland border and interior epithelial tissues.
(b) Prior USS methods in medical image research~\cite{SGSCN} and natural image research~\cite{PiCIE} vs. Our MSSG. \textcolor{mygland}{Green} and \textcolor{orange}{orange} regions denote the glands and the background respectively.}
\label{FigureOne}
\end{figure}

To reduce the annotation cost, developing annotation-efficient methods for semantic-level gland segmentation has attracted much attention~\cite{boxnet,OEEM,paul2016gland,egger2013pcg}. Recently, some researchers have explored weakly supervised semantic segmentation methods which use weak annotations (e.g., Bound Box~\cite{boxnet} and Patch Tag~\cite{OEEM}) instead of pixel-level annotations to train a gland segmentation network. However, these weak annotations are still laborious and require expert knowledge~\cite{boxnet}. 
To address this issue, previous works have exploited conventional clustering~\cite{paul2016gland,nguyen2010automated,4540990} and metric learning~\cite{tosun2010graph,egger2013pcg} to design annotation-free methods for gland segmentation. However, the performance of these methods can vary widely, especially in cases of malignancy.
This paper focuses on unsupervised gland segmentation, where \textbf{no annotations are required during training and inference.}

One potential solution is to adopt unsupervised semantic segmentation (USS) methods which have been successfully applied to medical image research and natural image research. On the one hand, existing USS methods have shown promising results in various medical modalities, e.g., magnetic resonance images~\cite{liu2020contrastive}, x-ray images~\cite{Aganj2018UnsupervisedMI,Huang2021ACM} and dermoscopic images~\cite{SGSCN}. However, directly utilizing these methods to segment glands could lead to over-segment results where a gland is segmented into many fractions rather than being considered as one target (see Fig.~\ref{FigureOne}(b)). This is because these methods are usually designed to be extremely sensitive to color~\cite{SGSCN}, while gland images present a unique challenge due to their highly dense and complex tissues with intricate color distribution~\cite{OEEM}. On the other hand, prior USS methods for natural images can be broadly categorized into coarse-to-fine-grained~\cite{SegSort,maskcontrast,EmergingPI,DSM,stego} and end-to-end (E2E) clustering ~\cite{deepcluster,IIC,PiCIE}. The former ones typically rely on pre-generated coarse masks (e.g., super-pixel proposals~\cite{SegSort}, salience masks~\cite{maskcontrast}, and self-attention maps~\cite{EmergingPI,DSM,stego}) as prior, which is not always feasible on gland images. The E2E clustering methods, however, produce under-segment results on gland images by confusing many gland regions with the background; see Fig.~\ref{FigureOne}(b). This is due to the fact that E2E clustering relies on the inherent connections between pixels of the same class~\cite{ssl_discriminative}, and essentially, grouping similar pixels and separate dissimilar ones. Nevertheless, the glands are composed of different parts (gland border and interior epithelial tissues, see Fig.~\ref{FigureOne}(a)) with significant variations in appearance. Gland borders typically consist of dark-colored cells, whereas the interior epithelial tissues contain cells with various color distributions that may closely resemble those non-glandular tissues in the background. As such, the E2E clustering methods tend to blindly cluster pixels with similar properties and confuse many gland regions with the background, leading to under-segment results.

To tackle the above challenges, \textbf{our solution is to incorporate an empirical cue about gland morphology as additional knowledge to guide gland segmentation}. The cue can be described as: \emph{Each gland is comprised of a border region with high gray levels that surrounds the interior epithelial tissues.}  
To this end, we propose a novel Morphology-inspired method via Selective Semantic Grouping, abbreviated as MSSG. To begin, we leverage the empirical cue to selectively mine out proposals for the two gland sub-regions with variant appearances. Then, considering that our segmentation target is the gland, we employ a Morphology-aware Semantic Grouping module to summarize the semantic information about glands by explicitly grouping the semantics of the sub-region proposals. In this way, we not only prioritize and dedicate extra attention to the target gland regions, thus avoiding under-segmentation; but also exploit the valuable morphology information hidden in the empirical cue, and force the segmentation network to recognize entire glands despite the excessive variance among the sub-regions, thus preventing over-segmentation. Ultimately, our method produces well-delineated and complete predictions; see Fig.~\ref{FigureOne}(b).

Our contributions are as follows: (1) We identify the major challenge encountered by prior unsupervised semantic segmentation (USS) methods when dealing with gland images, and propose a novel MSSG for unsupervised gland segmentation. (2) We propose to leverage an empirical cue to select gland sub-regions and explicitly group their semantics into a complete gland region, thus avoiding over-segmentation and under-segmentation in the segmentation results. (3) We validate the efficacy of our MSSG on two public glandular datasets (i.e., the GlaS dataset~\cite{glas} and the CRAG dataset~\cite{MIDL}), and the experiment results demonstrate the effectiveness of our MSSG in unsupervised gland segmentation.

\section{Methodology}
\begin{figure}[!t]
\centering
\includegraphics[width=\textwidth]{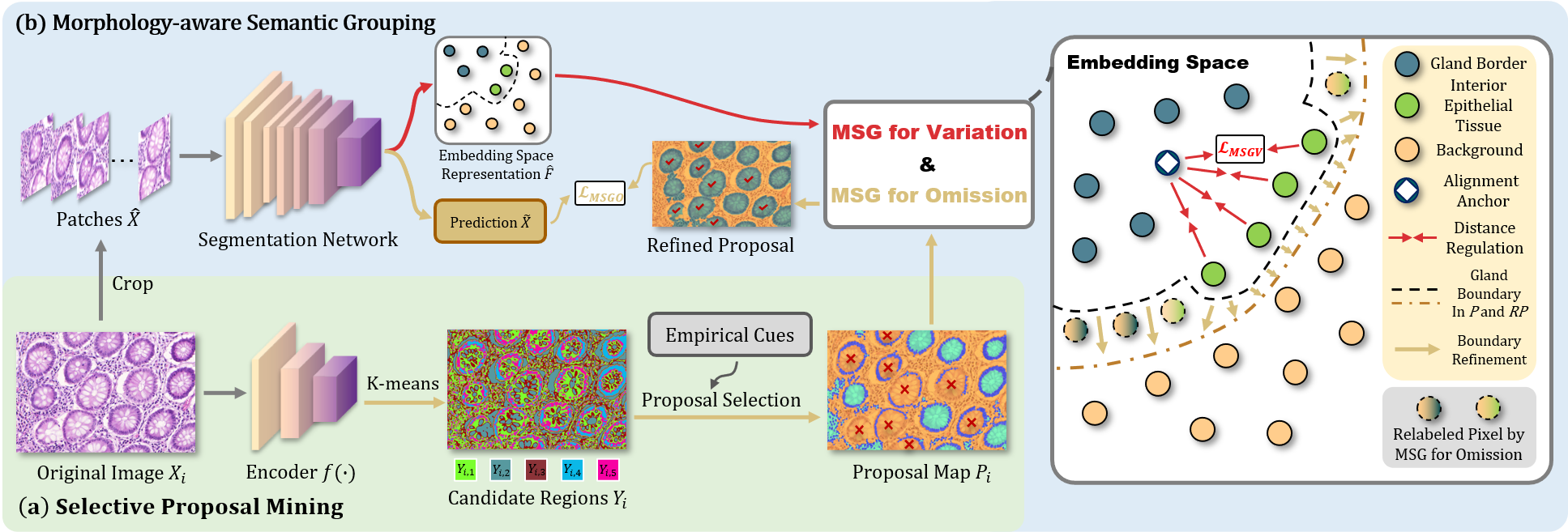}
\caption{Overview of our Morphology-inspired Unsupervised Gland Segmentation via Selective Semantic Grouping. (a) Selective Proposal Mining pipeline. We leverage an empirical cue to select proposals for gland sub-regions from the prediction of a shallow encoder $f(\cdot)$ which emphasizes low-level appearance features rather than high-level semantic features. (b) Morphology-aware Semantic Grouping (MSG) pipeline. We deploy a MSG for Variation module to group the two gland sub-regions in the embedding space with $L_{MSGV}$, and a MSG for Omission module to dynamically refine the proposal map generated by the proposal mining frame (see Gland boundary in $P$ and $RP$).} \label{FigureTwo}
\end{figure}

The overall pipeline of MSSG is illustrated in Fig.~\ref{FigureTwo}. The proposed method begins with a Selective Proposal Mining (SPM) module which generates a proposal map that highlights the gland sub-regions. The proposal map is then used to train a segmentation network. Meantime, a Morphology-aware Semantic Grouping (MSG) module is used to summarize the overall information about glands from their sub-region proposals. More details follow in the subsequent sections.
\subsection{Selective Proposal Mining}
Instead of generating pseudo-labels for the gland region directly from all the pixels of the gland images as previous works typically do, which could lead to over-segmentation and under-segmentation results, we propose using the empirical cue as extra hints to guide the proposal generation process.

Specifically, let the $i^{th}$ input image be denoted as $X_i \in \mathbb{R}^{C \times H \times W}$, where $H$, $W$, and $C$ refer to the height, width, and number of channels respectively. We first obtain a normalized feature map $F_{i}$ for $X_i$ from a shallow encoder $f$ with 3 convolutional layers, which can be expressed as $F_{i}=\left \| f\left(X_{i}\right) \right\|_2$. We train the encoder in a self-supervised manner, and the loss function $\mathcal{L}$ consists of a typical self-supervised loss $\mathcal{L}_{SS}$, which is the cross-entropy loss between the feature map $F_{i}$ and the one-hot cluster label $C_{i}=\arg\max \left(F_i\right)$, and a spatial continuity loss $\mathcal{L}_{SC}$, which regularizes the vertical and horizontal variance among pixels within a certain area $S$ to assure the continuity and completeness of the gland border regions \textcolor{red}{\textbf{(see Fig. 1 in the Supplementary Material)}}. The expressions for $\mathcal{L}_{SS}$ and $\mathcal{L}_{SC}$ are given below:
\begin{small}
\begin{equation}
    \mathcal{L}_{SS}(F_i[:, h, w], C_i[:, h, w])=-\mathop{\sum}_{d}^{D} C_{i}[d, h, w] \cdot \ln{F_i[d, h, w]}
    \label{equation1}
\end{equation}
\end{small}
\begin{footnotesize}
\begin{equation}
    \begin{split}
    \mathcal{L}_{SC} \left(F_{i}\right) = \sum_{\tiny s,h,w}^{\tiny S,H-s,W-s} \left(F_{i}[:, h+s, w]-F_{i}[:, h, w]\right)^2 \\
    + \left(F_{i}[:, h, w+s]-F_{i}[:, h, w]\right)^2.
    \end{split}
    \label{equation2}
\end{equation}
\end{footnotesize}

Then we employ K-means~\cite{kmean} to cluster the feature map $F_{i}$ into 5 candidate regions, denoted as $Y_{i}=\left\{y_{i,1}\in\mathbb{R}^{D \times n_0}, y_{i,2}\in\mathbb{R}^{D \times n_2}, ..., y_{i,5}\in\mathbb{R}^{D \times n_5} \right\}$, where $n_1+n_2+...+n_5$ equals the total number of pixels in the input image ($H\times W$).\\
\textbf{Sub-region Proposal Selection via the Empirical Cue.} The aforementioned empirical cue is used to select proposals for the gland border and interior epithelial tissues from the candidate regions $Y_{i}$. Particularly, we select the region with the highest average gray level as the proposal for the \emph{gland border}. Then, we fill the areas surrounded by the gland border proposal and consider them as the proposal for the \emph{interior epithelial tissues}, while the rest areas of the gland image are regarded as the background (i.e., non-glandular region). Finally, we obtain the proposal map $P_i \in \mathbb{R}^{3 \times H \times W}$, which contains the two proposals for two gland sub-regions and one background proposal.

\subsection{Morphology-aware Semantic Grouping}
A direct merge of the two sub-region proposals to train a fully-supervised segmentation network may not be optimal for our case. Firstly, the two gland sub-regions exhibit significant \textbf{variation} in appearance, which can impede the segmentation network's ability to recognize them as integral parts of the same object. Secondly, the SPM module may produce proposals with inadequate highlighting of many gland regions, particularly the interior epithelial tissues, as shown in Fig.~\ref{FigureTwo} where regions marked with \textcolor{red}{\bm{${\times}$}} are \textbf{omitted}. Consequently, applying pixel-level cross-entropy loss between the gland image and the merged proposal map could introduce undesired noise into the segmentation network, thus leading to under-segment predictions with confusion between the glands and the background. 
As such, we propose two types of Morphology-aware Semantic Grouping (MSG) modules (i.e., MSG for Variation and MSG for Omission) to respectively reduce the confusion caused by the two challenges mentioned above and improve the overall accuracy and comprehensiveness of the segmentation results. The details of the two MSG modules are described as follows.

Here, we first slice the gland image and its proposal map into patches as inputs. Let the input patch and its corresponding sliced proposal map be denoted as $\hat{X} \in \mathbb{R}^{C \times \hat{H} \times \hat{W}}$ and $\hat{P} \in \mathbb{R}^{3 \times \hat{H} \times \hat{W}}$. We can obtain the feature embedding map $\hat{F}$ which is derived as $\hat{F}=f_{feat}(\hat{X})$ and the prediction map $\widetilde{X}$ as $\widetilde{X}=f_{cls}(\hat{F})$, where $f_{feat}$ and $f_{cls}$ refers to the feature extractor and pixel-wise classifier of the segmentation network respectively.

\textbf{MSG for Variation} is designed to mitigate the adverse impact of appearance variation between the gland sub-regions. It regulates the pixel-level feature embeddings of the two sub-regions by explicitly reducing the distance between them in the embedding space. Specifically, according to the proposal map $\hat{P}$, we divide the pixel embeddings in $\hat{F} \in \mathbb{R}^{D \times \hat{H} \times \hat{W}}$ into gland border set $G=\left\{g_0, g_1,...,g_{k_g} \right\}$, interior epithelial tissue set $I=\left\{i_0, i_1,...,i_{k_i} \right\}$ and non-glandular (i.e., background) set $N=\left\{n_0, n_1,...,n_{k_n} \right\}$, where $k_g+k_i+k_n=\hat{H} \times \hat{W}$.
Then, we use the average of the pixel embeddings in gland border set $G$ as the alignment anchor and pull all pixels of $I$ towards the anchor:
\begin{small}
\begin{equation}
    \mathcal{L}_{MSGV} = \frac{1}{I} \mathop{\sum}_{i \in I} \left(i-\frac{1}{G} \sum_{g \in G} g \right)^2.
    \label{equation4}
\end{equation}
\end{small}

\textbf{MSG for Omission} is designed to overcome the problem of partial omission in the proposals. It identifies and relabels the overlooked gland regions in the proposal map and groups them back into the gland semantic category. To achieve this, for each pixel $n$ in the non-glandular (i.e., background) set $N$, two similarities are computed with the gland sub-regions $G$ and $I$ respectively:
\begin{small}
\begin{equation}
\begin{split}
    S_{n}^{G} = \frac{1}{G} \mathop{\sum}_{g \in G} \frac{g}{\left \| g \right\|_2} \cdot \frac{n}{\left \| n \right\|_2}, 
    S_{n}^{I} = \frac{1}{I} \mathop{\sum}_{i \in I} \frac{i}{\left \| i \right\|_2} \cdot \frac{n}{\left \| n \right\|_2}.
\end{split}
\label{equation5}
\end{equation}
\end{small}
$S_{n}^{G}$ (or $S_{n}^{I}$) represents the similarity between the background pixel $n$ and gland borders (or interior epithelial tissues). If either of them is higher than a preset threshold $\beta$ (set to 0.7), we consider $n$ as an overlooked pixel of gland borders (or interior epithelial tissues), and relabel $n$ to $G$ (or $I$). In this way, we could obtain a refined proposal map $RP$. Finally, we impose a pixel-level cross-entropy loss on the prediction and refined proposal $RP$ to train the segmentation network:
\begin{small}
\begin{equation}
    \mathcal{L}_{MSGO} = -\mathop{\sum}_{\hat{h},\hat{w}}^{\hat{H},\hat{W}} RP[:,\hat{h},\hat{w}]\cdot\ln{\widetilde{X}[:,\hat{h},\hat{w}]}, 
\label{equation6}
\end{equation}
\end{small}
The total objective function $\mathcal{L}$ for training the segmentation network can be summarized as follows:
\begin{small}
\begin{equation}
    \mathcal{L} = \mathcal{L}_{MSGO} + \lambda_{v}\mathcal{L}_{MSGV},
\label{equation7}
\end{equation}
\end{small}
where $\lambda_{v}$ (set to 1) is the coefficient.

\section{Experiments}
\subsection{Datasets}
We evaluate our MSSG on The Gland Segmentation Challenge (GlaS) dataset \cite{glas} and The Colorectal Adenocarcinoma Gland (CRAG) dataset \cite{MIDL}. The GlaS dataset contains 165 H\&E-stained histopathology patches extracted from 16 WSIs.
The CRAG dataset owns 213 H\&E-stained histopathology patches extracted from 38 WSIs. The CRAG dataset has more irregular malignant glands, which makes it more difficult than GlaS, and we would like to emphasize that the results on CRAG are from the model trained on GlaS without retraining.
\subsection{Implementation Details}
The experiments are conducted on four RTX 3090 GPUs. For the SPM, a 3-layer encoder is trained for each training sample. Each convolutional layer uses a $3\times3$ convolution with a stride of 1 and a padding size of 1. The encoder is trained for 50 iterations using an SGD optimizer with a polynomial decay policy and an initial learning rate of 1e-2. For the MSG, MMSegmentation~\cite{mmseg2020} is used to construct a PSPNet~\cite{PSPNet} with a ResNet-50 backbone as the segmentation network. The network is trained for 20 epochs with an SGD optimizer, a learning rate of 5e-3, and a batch size of 16. For a fair comparison, the results of all unsupervised methods in Tab.~\ref{table_1} are obtained using the same backbone trained with the corresponding pseudo-labels. The code is available at \url{https://github.com/xmed-lab/MSSG}.
\begin{figure*}[t]
	\centering
	\includegraphics[width=\linewidth,scale=1.0]{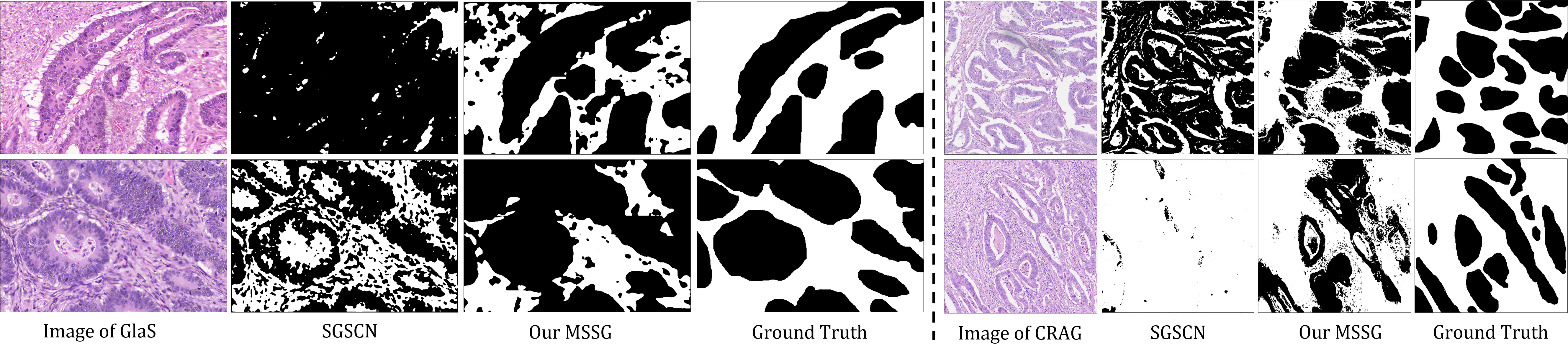}
	\caption{Visualization of predictions on GlaS (left) and CRAG dataset(right). Black denotes glandular tissues and white denotes non-glandular tissues \textcolor{red}{\textbf{(More in the Supplementary Material)}}.}
	\label{FigureThree}
\end{figure*}
\begin{table}[!t]
    \centering
    \caption{Comparison results on GlaS and CRAG dataset. \textbf{Bold} and \underline{underline} denote best and second-best results of the \emph{unsupervised methods}.}
\begin{tabular}{p{1.5cm}<{\centering}|p{2.5cm}<{\centering}|c|c|c c c}
    \hline
    Dataset & Method & Backbone & Supervision & F1 & DICE & mIOU                     \\ \hline
    \multirow{12}{*}{\makecell[c]{\textbf{GlaS}\\\textbf{Dataset}}} & Unet \cite{unet} & U-Net & Fully & 77.78\% & 79.04\% &65.34\%   \\
    & ResUNet \cite{resnet-unet} & U-Net & Fully & 78.83\% & 79.48\% & 65.95\%                                                  \\ 
    & MedT \cite{medT} & Transformer & Fully & 81.02\% & 82.08\% & 69.61\%                                                             \\ \cdashline{2-7}[2pt/2pt]
    & Randomly Initial & PSPNet & None & 49.72\% & 48.63\% & 32.13\%  \\
    & DeepCluster$^\ast$ \cite{deepcluster} & PSPNet & None & 57.03\% & 57.32\% & 40.17\% \\
    & PiCIE$^\ast$ \cite{PiCIE} & PSPNet & None & 64.98\% & 65.61\% & 48.77\%                                                     \\
    & DINO \cite{EmergingPI} & PSPNet & None & 56.93\% & 57.38\% &  40.23\%                                                       \\
    & DSM \cite{DSM} & PSPNet & None & 68.18\% & 66.92\% & 49.92\% \\
    & SGSCN$^\ast$ \cite{SGSCN} & PSPNet & None & \underline{67.62\%} & \underline{68.72\%} & \underline{52.16\%}                  \\
    & MSSG & PSPNet & None & \textbf{78.26\%} & \textbf{77.09\%} & \textbf{62.72\%} \\ 
    \hline
    \multirow{7}{*}{\makecell[c]{\textbf{CRAG}\\\textbf{Dataset}}} 
    & Unet \cite{unet} & U-Net & Fully & 82.70\% & 84.40\% & 70.21\%   \\
    & VF-CNN \cite{VF-CNN} & RotEqNet & Fully & 71.10\% & 72.10\% & 57.24\%  \\
    & MILDNet \cite{MIDL} & MILD-Net & Fully & 86.90\% & 88.30\% & 76.95\% \\ \cdashline{2-7}[2pt/2pt]
    & PiCIE$^\ast$ \cite{PiCIE} & PSPNet & None & 67.04\% & 64.33\% & 52.06\%  \\
    & DSM \cite{DSM} & PSPNet & None & 67.22\% & 66.07\% & 52.28\% \\
    & SGSCN$^\ast$ \cite{SGSCN} & PSPNet & None & \underline{69.29\%} & \underline{67.88\%} & \underline{55.31\%} \\
    & MSSG & PSPNet & None & \textbf{77.43\%} & \textbf{77.26\%} & \textbf{65.89\%} \\ 
    \hline
\end{tabular}
	\label{table_1}
\end{table}
\subsection{Comparison with state-of-the-art methods}
We compare our MSSG with multiple approaches with different supervision settings in Tab. \ref{table_1}. On the GlaS dataset, the end-to-end clustering methods (denoted by “$\ast$”) end up with limited improvement over a randomly initialized network. Our MSSG, on the contrary, achieves significant advances. Moreover, MSSG surpasses all other unsupervised counterparts, with a huge margin of 10.56\% at mIOU, compared with the second-best unsupervised counterpart. On CRAG dataset, even in the absence of any hints, MSSG still outperforms all unsupervised methods and even some of the fully-supervised methods. Additionally, we visualize the segmentation results of MSSG and its counterpart (i.e., SGSCN~\cite{SGSCN}) in Fig. \ref{FigureThree}. On both datasets, MSSG obtains more accurate and complete results.
\begin{figure*}[t]
	\centering
	\includegraphics[width=\linewidth,scale=1.0]{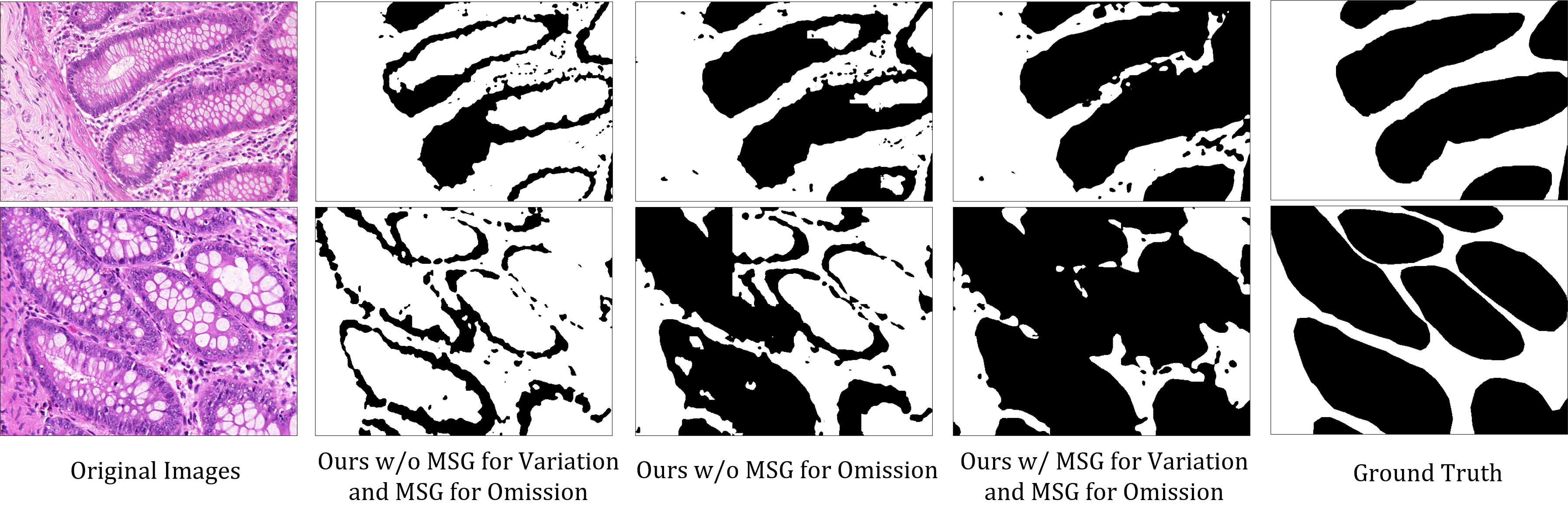}
	\caption{Ablation Study on MSG modules. Without MSG, the performance is not good enough, due to \emph{significant sub-region variation} and \emph{gland omission}. With MSG modules, the performance of the network is progressively improved \textcolor{red}{\textbf{(More in the Supplementary Material)}}.}
	\label{FigureFour}
\end{figure*}
\begin{table}[t]
	\centering
	\caption{Performance gains with MSG modules. The segmentation performance is progressively improved as the involvement of MSG for Variation \& MSG for Omission.}
	\begin{tabular}{p{3cm}<{\centering} p{3cm}<{\centering}|p{1.5cm}<{\centering} p{3cm}<{\centering}}
		\hline
		MSG for Variation & MSG for Omission & mIOU & \textbf{\textcolor{teal}{Improvement($\Delta$)}} \\ \hline
		\bm{${\times}$} & \bm{${\times}$} & 48.42\% & \textbf{\textcolor{teal}{-}}                       \\
		\bm{${\surd}$} & \bm{${\times}$} & 56.12\% & \textbf{\textcolor{teal}{+7.70\%}}                     \\
		\bm{${\times}$} & \bm{${\surd}$} & 50.18\% & \textbf{\textcolor{teal}{+1.64\%}}                     \\
		\bm{${\surd}$} & \bm{${\surd}$} & 62.72\% & \textbf{\textcolor{teal}{+14.30\%}}                    \\ \hline
	\end{tabular}
	\label{table_4}
\end{table}
\subsection{Ablation Study}
Tab. \ref{table_4} presents the ablation test results of the two MSG modules. It can be observed that the segmentation performance without the MSG modules is not satisfactory due to the \emph{significant sub-region variation} and \emph{gland omission}. With the gradual inclusion of the MSG for Variation and Omission, the mIOU is improved by 6.42\% and 2.57\%, respectively. Moreover, with both MSG modules incorporated, the performance significantly improves to 62.72\% (+14.30\%). 
we also visualize the results with and without MSG modules in Fig.~\ref{FigureFour}. It is apparent that the model without MSG ignores most of the interior epithelial tissues. With the incorporation of MSG for Variation, the latent distance between gland borders and interior epithelial tissues is becoming closer, while both of these two sub-regions are further away from the background. As a result, the model can highlight most of the gland borders and interior epithelial tissues. Finally, with both MSG modules, the model presents the most accurate and similar result to the ground truth.
\textbf{\textcolor{red}{More ablation tests on the SPM (Tab. 1 \& 2) and hyper-parameters (Tab. 3) are in the Supplementary Material.}}

\section{Conclusion}
This paper explores a DL method for unsupervised gland segmentation, which aims to address the issues of over/under segmentation commonly observed in previous USS methods. The proposed method, termed MSSG, takes advantage of an empirical cue to select gland sub-region proposals with varying appearances. Then, a Morphology-aware Semantic Grouping is deployed to integrate the gland information by explicitly grouping the semantics of the selected proposals. By doing so, the final network is able to obtain comprehensive knowledge about glands and produce well-delineated and complete predictions. Experimental results prove the superiority of our method qualitatively and quantitatively. 

\subsubsection{Acknowledgement.} 
This work was supported in part by the Hong Kong Innovation and Technology Fund under Project ITS/030/21 and in part by a grant from Foshan HKUST Projects under FSUST21-HKUST11E. 

\bibliographystyle{splncs04}
\bibliography{paper1196}
\end{document}